\documentclass[12pt]{article}
\usepackage{amsmath}
\usepackage{times}
\usepackage{graphicx}
\usepackage[usenames,dvipsnames]{xcolor}
\usepackage{multirow}
\usepackage[authoryear]{natbib}
\usepackage{rotating}
\usepackage{bbm}
\usepackage{latexsym}
\usepackage{hyperref}

\usepackage{amssymb, amsmath}
\usepackage{graphicx, subfigure}
\newcommand{\mb}{\mathbf}

\newcommand{\argmin}{\operatornamewithlimits{argmin}}

\newcommand{\Real}{{\mathcal R}}
\newcommand{\Imag}{{\mathcal I}}
\newcommand{\Complex}{{\mathcal C}}

\newcommand{\captionfonts}{\normalsize}

\makeatletter  
\long\def\@makecaption#1#2{%
  \vskip\abovecaptionskip
  \sbox\@tempboxa{{\captionfonts #1: #2}}%
  \ifdim \wd\@tempboxa >\hsize
    {\captionfonts #1: #2\par}
  \else
    \hbox to\hsize{\hfil\box\@tempboxa\hfil}%
  \fi
  \vskip\belowcaptionskip}
\makeatother   

\begin{document}
\hspace{13.9cm}1

\ \vspace{20mm}\\

{\LARGE An unsupervised algorithm for learning Lie group transformations}

\ \\
{\bf \large Jascha Sohl-Dickstein$^{\displaystyle 1, \displaystyle 4, \displaystyle 5}$}\\
{\bf \large Ching Ming Wang$^{\displaystyle 1, \displaystyle 2, \displaystyle 6}$}\\
{\bf \large Bruno Olshausen$^{\displaystyle 1, \displaystyle 2, \displaystyle 3}$}\\
{$^{\displaystyle 1}$Redwood Center for Theoretical Neuroscience, $^{\displaystyle 2}$School of Optometry, $^{\displaystyle 3}$Helen Wills Neuroscience Institute, {University of California, Berkeley.}\\
{$^{\displaystyle 4}$Department of Applied Physics, Stanford University.}\\
{$^{\displaystyle 5}$Khan Academy.}\\
{$^{\displaystyle 6}$earthmine inc.}\\
%


\thispagestyle{empty}
\markboth{}{NC instructions}
\ \vspace{-0mm}\\
%
\begin{center} {\bf Abstract} \end{center}

We present several theoretical contributions which allow Lie groups to be fit to high dimensional datasets. Transformation operators are represented in their eigen-basis, reducing the computational complexity of parameter estimation to that of training a linear transformation model. A transformation specific ``blurring" operator is introduced that allows inference to escape local minima via a smoothing of the transformation space. A penalty on traversed manifold distance is added which encourages the discovery of sparse, minimal distance, transformations between states. Both learning and inference are demonstrated using these methods for the full set of affine transformations on natural image patches.  Transformation operators are then trained on natural video sequences. It is shown that the learned video transformations provide a better description of inter-frame differences than the standard motion model based on rigid translation. 



\section{Introduction}

Over the past several decades, research in the natural scene statistics community has shown that it is possible to learn efficient representations of sensory data, such as images and sound, from their intrinsic statistical structure. Such representations exhibit higher coding efficiency as compared to standard Fourier or wavelet bases \citep{Lewicki99}, and they also match the neural representations found in visual and auditory systems \citep{Olshausen1996,hateren_schaaf_1998,smith2006efficient}.  Here we explore whether such an approach may be used to learn {\em transformations} in data by observing how patterns change over time, and we apply this to the problem of coding image sequences, or video. 

A central problem in video coding is to find representations of image sequences that efficiently capture how image content changes over time.  Current approaches to this problem are largely based on the assumption that local image regions undergo a rigid spatial translation from one frame to the next, and they encode the local motion along with the resulting prediction error \citep{Wiegand03}. However, the spatiotemporal structure occurring in natural image sequences can be quite complex due to occlusions, the motion of non-rigid objects, lighting and shading changes, and the projection of 3D motion onto the 2D image plane. Attempting to capture such complexities with a simple translation model leads to larger prediction errors and thus a higher bit rate. 
The right approach would seem to require a more complex transformation model that is adapted to the statistics of how images actually change over time.

The approach we propose here is based on learning a Lie (continuous transformation) group representation of the dynamics which occur in the visual world \citep{Rao99,Miao07,Culpepper10,cohen2014learning}.  
The Lie group is built
by first describing each of the infinitesimal transformations which an image may
undergo. The full group is then generated from all possible compositions of
those infinitesimal transformations, which allows for transformations to be
applied smoothly and continuously. A large class of
visual transformations, including all the affine transformations, intensity
changes due to changes in lighting, and contrast changes can be described simply using Lie group
operators.  Spatially localized versions of the preceding transformations can also be captured. 
In \citep{Miao07,Rao99}, Lie group operators were trained on
image sequences containing a subset of affine transformations.  \citep{Memisevic10} trained a second order restricted Boltzmann machine on pairs of frames, an alternative technique which also shows promise for capturing temporal structure in video.

Despite the simplicity and power of the Lie group
representation, training such a model is difficult, in part due to the
high computational cost of evaluating and propagating 
learning gradients through matrix exponentials.
Previous work \citep{Rao99,Miao07,Olshausen07} has approximated the full model using a first order Taylor
expansion, reducing the exponential model to a linear one. 
While computationally efficient, a linear model approximates the full exponential model only for a small range of transformations.  This can be a hinderance in dealing with real video data, which can contain large changes between pairs of frames.  
Note that in \citep{Miao07}, while the full Lie group model is used in inferring transformation coefficients, only its linear approximation is used during learning.
\citep{Culpepper10} work with a full exponential model, but that technique requires performing a costly eigendecomposition of the effective transformation operator for each sample and at every learning or inference step.

Another hurdle one encounters in using a lie group model to describe transformations is that the inference process, which computes
the transformation between a pair of images, is highly non-convex with many
local minima. This problem has been
extensively studied in image registration, stereo matching and the computation
of optic flow. For a specific set of transformations (translation, rotation
and isotropic scaling), \citep{Kokiopoulou09} showed that one could find the global minimum by
formulating the problem using an overcomplete image representation composed of Gabor and Gaussian basis functions.
For arbitrary transformations, one solution is to initialize inference with many different
coefficient values \citep{Miao07}; but the drawback here is that the number of initial guesses 
grows exponentially with the number of transformations. Alternatively,
\citep{Lucas81,Vasconcelos97,Black96} perform matching
with an image pyramid, using solutions from a lower
resolution level to seed the search algorithm at a higher resolution level.
\citep{Culpepper10} used the same technique to perform learning with Lie Group
operators on natural movies.
Such piecewise coarse-to-fine schemes avoid local minima by searching in the smooth parts of the transformation space before proceeding to less smooth parts of the space.  This constitutes an indirect method of coarsening the transformation through spatial blurring.  As we show here, it is also possible to smooth the transformation space directly, resulting in a robust method for estimating transformation coefficients for arbitrary transformations.

In this work we propose a method for directly learning the Lie group operators that mediate continuous transformations in natural movies, and we demonstrate the ability to robustly infer transformations between frames of video using the learned operators.  The computational complexity of learning the operators is reduced by re-parametrizing them in terms of their eigenvectors and eigenvalues, resulting in a complexity equivalent to that of the linear approximation.  Inference is made robust and tractable by smoothing the transformation space directly, which allows for a continuous coarse-to-fine search for the transformation coefficients.  Both learning and inference are demonstrated on test data containing a full set of known affine transformations.  
The same technique is then used to learn a set of
Lie operators describing changes between frames in natural movies, where the optimal solution is not known.
Unlike previous Lie group implementations, 
we demonstrate an ability to work
simultaneously with multiple transformations and large inter-frame differences
during both inference and learning. 
In another paper we additionally show the utility of this approach for video compression \citep{DCC_11}.

\section{Model}

As in \citep{Rao99}, we consider the class of continuous transformations described by the first order differential equation
\begin{eqnarray}
	\frac
		{\partial \mb x\left( \mu \right )}
		{\partial \mu}
	& = &
		A \ \mb x\left( \mu \right )
		,
  \label{equ:derivative}
\end{eqnarray}
where $\mb x\left( \mu \right ) \in \Real^{N \times 1}$ represents the pixel values in a $\sqrt{N} \times \sqrt{N}$ image patch, $\mb A \in \Real^{N \times N}$ is an infinitesimal transformation operator and the generator of the Lie group, and $\mb x\left( \mu \right ) \in \Real^{N \times 1}$ is the image $\mb x\left(0\right)$
transformed by $\mb A$ by an amount $\mu \in \Real$.  This differential equation has solution
\begin{equation}
 	\mb x\left( \mu \right) = e^{\mb A \mu} \mb x\left(0\right) = T\left( \mu \right)
\mb x\left(0\right)
.
	\label{equ:deriv_solution}
\end{equation}
$T\left( \mu \right) = e^{\mb A \mu}$ is a matrix exponential defined by its Taylor expansion.
 
Our goal is to use transformations of this form to model the
changes between adjacent frames $\mb x^{(t)}, \mb x^{(t+1)} \in \Real^{N \times 1}$ that occur in
natural video.  
That is, we seek to find the model parameters $A$ (adapted over an ensemble of video image sequences) and coefficients $\mu^{(t)}$ (inferred for each pair of frames) that minimize the reconstruction error
\begin{equation}
	E
		=
	\sum_t
	\left| \left|
		\mb x^{(t+1)} - T\left( \mu^{(t)} \right) \mb x^{(t)}
	\right| \right|_2^2
	\label{equ:err}
.
\end{equation}
This will be extended to multiple transformations below.


\subsection{Eigen-decomposition}

To derive a learning rule for $\mb A$, it is necessary to compute the
gradient $\frac{\partial e^{\mb A\mu} }{\partial \mb A}$.  Under naive application of the chain rule this requires $O\left( N^6
\right)$ operations \citep{Ortiz01} ($O\left(N^2\right)$ operations per element $\times$ $N^4$ elements), making it computationally intractable for many problems of
interest.  However, this computation reduces to $\approx O\left( N^2.4 \right)$ (the complexity of matrix inversion) if $A$ is rewritten in terms of its
eigen-decomposition,
	$\mb A = \mb U \mb \Lambda \mb U^{-1}$,
and learning is instead performed directly in terms of $\mb U$ and
$\mb \Lambda$.  $\mb U \in \Complex^{N \times N}$ is a complex matrix consisting of the eigenvectors of $\mb A$,
$\mb \Lambda \in \Complex^{N \times N}$ is a complex diagonal matrix holding the eigenvalues of $\mb A$, and $\mb U^{-1}$
is the inverse of $\mb U$. The matrices must be complex in order to
facilitate periodic transformations, such as rotation.   Note that $\mb U$ need not be orthonormal.  The benefit of this
representation is that 
\begin{equation}
	e^{\mb U \mb \Lambda \mb U^{-1} \mu} = \mb I + \mb U \mb \Lambda \mb U^{-1} \mu + \frac{1}{2}\mb U \mb \Lambda
\textcolor{Gray}{\mb U^{-1} \mb U} \mb \Lambda \mb U^{-1} \mu^2 + \ldots
	= \mb U e^{\mu\mb \Lambda} \mb U^{-1}
\end{equation}
where the matrix exponential of a diagonal matrix is simply the element-wise
exponential of its diagonal entries. This representation therefore replaces the
full matrix exponential by two matrix multiplications, a matrix inversion, and an element-wise
exponential.

This change of form enforces the restriction that $\mb A$
be diagonalizable. We do not expect this restriction to be onerous, since the set of 
non-diagonalizable matrices is measure zero in the set of all matrices. 
However some specific desirable transformations may be lost. 
For instance, since nilpotent matrices are non-diagonalizable, translation without periodic boundary conditions (translation where content is lost as it moves out of frame) cannot be captured by this approach.

\subsection{Adaptive Smoothing}

In general the reconstruction error described by Equation
\ref{equ:err} is highly non-convex in $\mu$ and contains many local minima. To
illustrate, the red solid line in Figure \ref{fig:blur} plots the reconstruction
error for a
white-noise image patch shifted by three pixels to the right as a function of
transformation coefficient $\mu$ for a generator $\mb A$ corresponding to left-right
translation.  It is clear that performing gradient descent-based inference on $\mu$ for
this error function would be problematic.

To overcome this problem, we propose a multi-scale search 
that adaptively smooths the
operator over a range of transformation coefficients. 
Specifically, the operator is averaged 
over a range of transformations using a Gaussian weighting for the
coefficient values
\begin{eqnarray}
	T\left(\mu, \sigma \right) & = & \int_{-\infty}^\infty T\left( s \right)
		\frac{ 1	}{\sqrt{2\pi} \sigma} 
			e^{-\frac{\left(s - \mu\right)^2}{2 \sigma^2}} ds \\
	& = & \mb U \left[ \int_{-\infty}^\infty e^{\mb \Lambda s - \mb I \frac{\left(s - \mu\right)^2}{2 \sigma^2}} 
		\frac{ 1	}{\sqrt{2\pi} \sigma} 
		 ds \right] \mb U^{-1} \\
	& = &
		\mb U e^{\mu \Lambda} e^{\frac{1}{2} \mb \Lambda^2 \sigma^2}\mb U^{-1}
	.
\end{eqnarray}
$T\left(\mu, \sigma \right)$ replaces $T\left( \mu \right)$ in Equation~\ref{equ:err}.  The error is then minimized with respect to both $\mu$ and $\sigma$.\footnote{This can alternatively be seen as introducing an additional
transformation operator, this one a smoothing operator
\begin{eqnarray*}
\mb A_{smooth} & = & \frac{1}{2} \mb U \mb \Lambda^2 \mb U^{-1},
\end{eqnarray*}
with coefficient $\mu_{smooth} = \sigma^2$.
$\mb A_{smooth}$ smooths along the transformation direction given by $\mb A = \mb U \mb \Lambda \mb U^{-1}$.}

Instead of finding the single best value of $\mu$ that minimizes $E$, 
$T\left(\mu, \sigma \right)$ allows for a Gaussian distribution over $\mu$, effectively blurring the signal along the transformation direction given by $\mb A = \mb U \mb \Lambda \mb U^{-1}$. In
the case of translation, for instance, this averaging over a range of
transformations blurs the image along the direction of translation.  
The higher the value of $\sigma$, the larger the blur.  This blurring is specific to the operator $\mb A$ (e.g., if $\mb A$ corresponded to rotation in one quadrant of an image patch, then this averaging would lead to rotational blur in that single quadrant).
Under simultaneous inference in $\mu$
and $\sigma$, images are matched first at a coarse scale, and the match
refines as the blurring of the image decreases. 
This approach is in the spirit of 
Arathorn's map-seeking circuit \citep{Arathorn02}, 
and also in the spirit of the transformation specific blurring kernels 
presented in \citep{Mobahi2012}.  
\citep{Mobahi2012} differs in that they 
fix the blurring coefficient $\sigma$ rather than inferring it; in that 
their smoothing is specific to 
affine transformations and homography, while ours applies to all 
first order differential operators; 
and that they 
smooth the full objective function, 
rather than the 
transformation kernel.

To illustrate the way in which the proposed transformation leads to better
inference, the dotted lines in Figure \ref{fig:blur} shows the reconstruction
error as a function of $\mu$ for different values of $\sigma$. Note that,
by allowing $\sigma$ to vary, steepest
descent paths open out of the local minima, detouring through coarser scales.
\begin{figure}
\begin{center}
	\includegraphics[width=10cm]{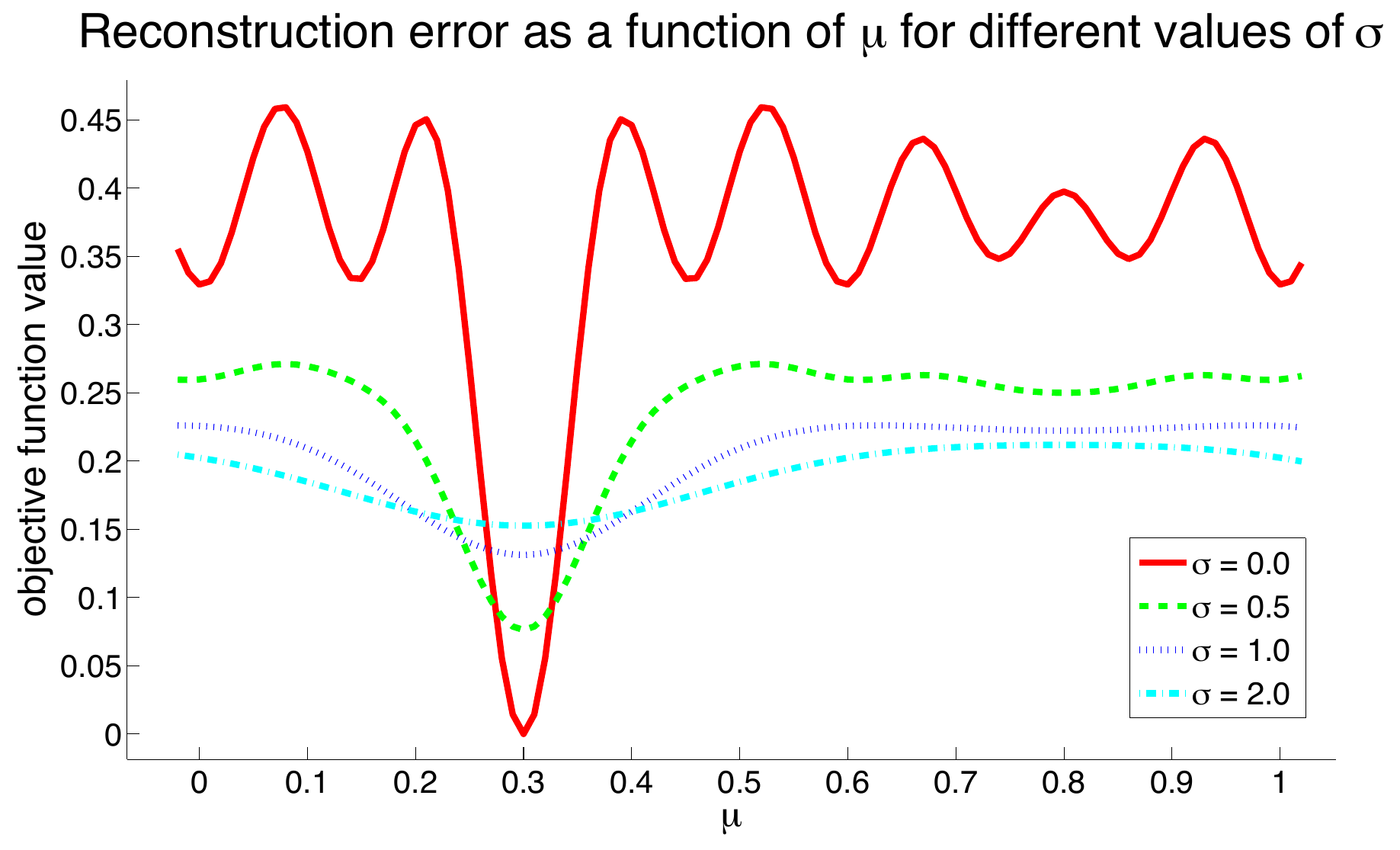}
\end{center}
\caption{
Local minima in the error function landscape can be escaped by increasing the smoothing coefficient $\sigma$.  
This plot shows reconstruction error (Equation
\ref{equ:err}) as a function of transformation coefficient $\mu$ for several values of the smoothing coefficient $\sigma$.  In this case the target pattern $x^{(t+1)}$ has been translated in one dimension relative to an initial white noise pattern $x^{(t)}$, and the operator $A$ is the one-dimensional translation operator. 
}
\label{fig:blur}
\end{figure}

\subsection{Multiple Transformations}

A single transformation is inadequate to describe most changes observed in the
visual world.  The model presented above can be extended to multiple
transformations by concatenating transformations in the following way:
\begin{eqnarray}
\label{eqn:multi}
		T_{\mathrm{multi}}\left( \mb \mu, \mb \sigma \right)
	& = &
		T_1\left( \mu_{1}, \sigma_{1} \right) T_2\left( \mu_{2},
\sigma_{2} \right)...
	=
		\prod_k
			T_k\left( \mu_{k}, \sigma_{k} \right) \\
	T_k\left( \mu_{k}, \sigma_{k} \right) & = & \mb U_k e^{\mu_{k} \mb \Lambda_k}
e^{\frac{1}{2} \mb \Lambda_k^2 \sigma_{k}^2}\mb U_k^{-1}
\end{eqnarray}
where $k$ indexes the transformation. Note that the transformations $T_k\left(
\mu_{k}, \sigma_{k} \right)$ do not in general commute, and thus the ordering
of the terms in the product must be maintained.

Because of the fixed ordering of transformations and due to the lack of
commutativity, the multiple transformation case no longer constitutes a
Lie group for most choices of transformation generators $\mb A$.  Describing the group
structure of this new model is a goal of future work.  For the present, we
note that many obvious video transformations -- affine transformations, brightness
scaling, and contrast scaling -- can be fully captured by the model form in Equation
\ref{eqn:multi}, though the choice of coefficient values $\mu_{k}$ for a transformation may depend 
on the order of terms in the product.

\subsection{Regularization via Manifold Distance}
In order to encourage the learned operators to act independently of each other,
and to learn to transform between patches in the most direct way possible, we
penalize the distance through which the transformations move the image through pixel space. 
Since this penalty consists of a sum of the distances traversed by each operator, it acts similarly to an L1 penalty in sparse coding, and encourages travel between two points to occur via 
a path described by a single transformation, rather than by a longer path described by multiple transformations.  

The distance traversed by the transformations can be expressed as
\begin{eqnarray}
d_{\mathrm{multi}} = \sum_k d_k \left( \mu_{k}, \mb y_k(0) \right)
\end{eqnarray}
where $\mb y_k(0) = \prod_{m<k} T_m\left(
\mu_{m}, \sigma_{m} \right) \mb x(0)$ is the image patch prior to application of
transformation $k$. Assuming a Euclidean metric, and neglecting the effect of adaptive blurring, the distance $d_k\left(  \mu_{k}, \mb y_k(0) \right)$ traversed by each single transformation in
the chain is
\begin{eqnarray}
d_k\left( \mu_{k}, \mb y_k(0) \right)
 & = & \int_{0}^{\mu_k} || \dot{\mb y}_k(\tau) ||_2 d\tau\\	
  & = & \int_{0}^{\mu_k} || \mb A_k \mb y_k(\tau) ||_2 d\tau\\
  & = & \int_{0}^{\mu_k} || \mb A_k e^{\mb A_k \tau} \mb y_k(0) ||_2 d\tau
\end{eqnarray}
Finding a closed form solution for the above integral is difficult, but
it can be approximated using a linearization around $\tau = \frac{\mu_k}{2}$,
\begin{equation}
 d_k\left(  \mu_{k}, \mb y_k(0) \right) \approx \left| \mu_{k} \right| || \mb A_k e^{\mb A_k \frac{\mu_{k}}{2}} \mb y_k(0) ||_2^1
 .
\end{equation}


\subsection{The Complete Model}

Putting together the components above, the full model's objective function is
\begin{equation}
 \begin{split}
  	E\left(\mb \mu,\mb \sigma, \mb U, \mb \Lambda, \mb x \right) = \ \ \ \ \ & \eta_n \sum_t
	\left| \left|
		\mb x^{(t+1)} - T_{multi}\left( \mb \mu^{(t)}, \mb \sigma^{(t)}  \right) \mb x^{(t)}
	\right| \right|_2^2 \\
	+ \ & \eta_d \sum_t \sum_k \mu_{k}^{(t)} ||
\mb A_k e^{\frac{\mu_{k}^{(t)}}{2}\mb A_k } \mb x^{(t)}_k ||_2^1 \\
	+ \ & \eta_\sigma \sum_t \sum_k (\sigma_{k}^{(t)})^2
 \end{split}
 \label{equ:E full}
\end{equation}
A small L2 regularization term on $\sigma_{k}^{(t)}$ is included as it was found to speed convergence during learning.  We used $\eta_n = 1$, $\eta_d = 0.005$, and $\eta_\sigma = 0.01$.


\subsection{Inference and Learning}

To find \textbf{U} and $\mathbf{\Lambda}$, we use 
Expectation-Maximization with a MAP approximation to the expectation step. This iterates between
the following two steps:
\begin{enumerate}
 \item For a set of video patches $\mb x$, find optimal estimates for the latent variables $\hat{\mb \mu}$ and $\hat{\mb \sigma}$ for each pair of frames, while holding the estimates of the model parameters $\hat{\mb U}$ and $\hat{\mb \Lambda}$ fixed,
\begin{equation}
\hat{\mb \mu}, \hat{\mb \sigma} = \argmin_{\mb \mu, \mb \sigma} E\left(\mb \mu,\mb \sigma, \hat{\mb U}, \hat{\mb \Lambda}, \mb x \right)
.
 \label{equ:inference}
\end{equation}
 \item Optimize the estimated model parameters $\hat{\mb U}$ and $\hat{\mb \Lambda}$ while holding the latent variables fixed,
\begin{equation}
\hat{\mb U}, \hat{\mb \Lambda} = \argmin_{\mb U, \mb \Lambda} E\left(\hat{\mb \mu}, \hat{\mb \sigma}, \mb U, \mb \Lambda, \mb x \right)
 \label{equ:learning}
\end{equation}
\end{enumerate}
All optimization was performed using the L-BFGS implementation in minFunc \citep{Schmidt05}.  

Note that there is a degeneracy in $\mb U_k$, due to the fact that the columns (corresponding to the
eigenvectors of $\mb A_k$) can be rescaled arbitrarily, and $\mb A_k$ will remain
unchanged as the inverse scaling will occur in the rows of $\mb U_k^{-1}$.  If not dealt with, $\mb U_k$ and/or $\mb U_k^{-1}$ can random walk in column or row length over many learning steps, until one or both is ill conditioned.  As described in detail in the appendix, this effect is compensated for by rescaling the columns of $\mb U_k$ such that they have identical power to the corresponding rows in $\mb U_k^{-1}$.  Similarly, there is a degeneracy in the relative scaling of $\mb \mu$ and $\mb \Lambda$.  In order to prevent $\hat{\mb\mu}$ or $\hat{\mb\Lambda}$ from random walking to large values, after every iteration $\hat\mu_k$ is multiplied by a scalar $\alpha_k$, and $\hat{\mb\Lambda}_k$ is divided by $\alpha_k$, where $\alpha_k$ is chosen such that $\hat{\mb\mu_k}$ has unit variance over the training set.

\section{Experimental Results}
\subsection{Inference with Affine Transforms}\label{section:infer} 
To verify the correctness of the proposed inference algorithm a test dataset containing a set of known transformations was constructed by applying affine transformations to natural image patches.  The transformation coefficients were then inferred using a set of pre-specified operators matching those used to construct the dataset.  
For this purpose a pool of 1000 $11 \times 11$ natural image patches were cropped from a set of short video clips
from The BBC's \emph{Animal World Series}.  
Each image patch was transformed
by the full set of affine transformations simultaneously with the transformation
coefficients drawn uniformly from the ranges listed below. \footnote{Vertical
skew is left out since it can be constructed using a combination of the other
affine operators.}
\begin{center}
\begin{tabular}{|c|c|}
 \hline 
    Transformation Type & Range \\
	\hline
    horizontal translation & $\pm$ 5 pixels \\	
    vertical translation & $\pm$ 5 pixels \\	
    rotation & $\pm$ 180 degrees \\	
    horizontal scaling & $\pm$ 50 \% \\	
    vertical scaling & $\pm$ 50 \% \\	
    horizontal skew & $\pm$ 50 \% \\	
  \hline
\end{tabular}
\end{center}

The eigenvector matrices $\mb U_k$ were initialized with isotropic unit norm zero mean Gaussian noise, followed by the transformation described in Appendix \ref{u_rescale}. 
The eigenvalues $\Lambda_{k}$ were initialized as $\Lambda_{kjj} = \tilde{n}_{\mathcal I} i + 0.01 \tilde{n}_{\mathcal R}$, where $\Lambda_{kjj}$ is the $j$th diagonal entry in eigenvalue matrix $\Lambda_{k}$, 
$\tilde{n}_{\mathcal I}$ and $\tilde{n}_{\mathcal R}$ are both unit norm zero mean Gaussians, and $i = \sqrt{-1}$. Finally, for each sample $\mu_k$ and $\sigma_k$ were initialized as 
$\mu_k \sim \mathcal N\left(0, 10^{-4} \right), \sigma_k \sim \mathcal N\left(0, 0.04 \right)$.

The proposed inference algorithm (Equation \ref{equ:inference}) is used to recover the transformation
coefficients. Figure \ref{fig:SNR} shows the fraction of the recovered coefficients which differed by less than 1\%
from the true coefficients. The distribution
of the PSNR of the reconstruction is also shown. The
inference algorithm recovers the coefficients with a high degree of accuracy.  The PSNR in the
reconstructed images patches was higher than 25dB for 85\% of the
transformed image patches. In addition, we found that adaptive blurring
significantly improved inference, as evident in Figure \ref{fig:SNR}a.

\begin{figure}
\begin{center}
(a)  \includegraphics[height=5.8cm,width=6.5cm]{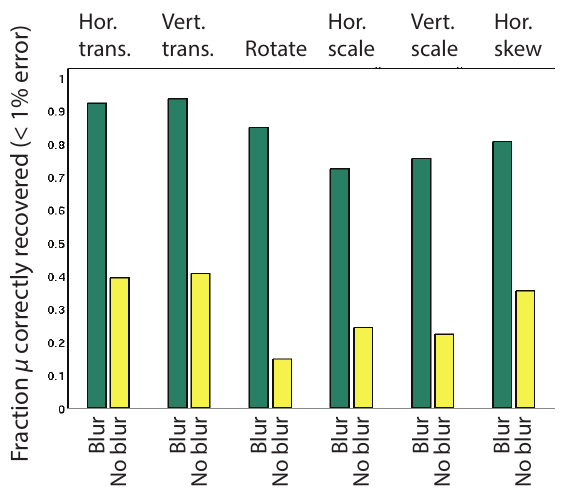}
(b)  \includegraphics[height=5.8cm,width=6.5cm]{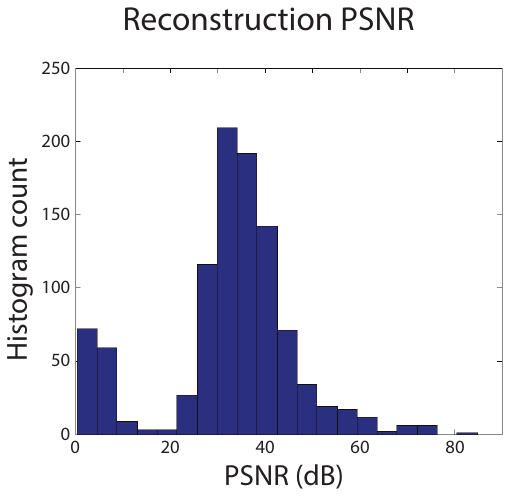}  
\end{center}
\caption{ 
Performance in inferring transformation coefficients. 
{\em(a)} The fraction of recovered coefficients which differed by
less than 1\% from the true coefficient values.  Image patches were transformed using a set
of hand coded affine transformations (all transformations simultaneously), and recovery was performed via gradient descent of Equation \ref{equ:E full}.  Inference with (``blur'') and without (``no blur'') 
adaptive smoothing is compared. {\em(b)} The distribution of
PSNR values for image patches reconstructed using coefficients inferred
with adaptive smoothing.  The majority of transformed image patches are reconstructed with high PSNR. }
\label{fig:SNR}
\end{figure}

\subsection{Learning Affine Transformations} \label{sec affine learn}
To demonstrate the ability to learn transformations, we trained the algorithm on image sequences transformed by a single affine transformation operator (translation, rotation, scaling, or skew).
 The training data used 
 were single image patches from the same BBC clips as in Section \ref{section:infer},
 transformed by an open source Matlab package \citep{Shen08} with the same transformation range used in Section \ref{section:infer}. 
 Initialization was the same as in Section \ref{section:infer}.

The affine transformation operators are spatial derivative operators in
the direction of motion. For example, a horizontal translation operator is a
derivative operator in the horizontal direction while a rotation operator
computes a derivative circularly. Our learned operators exhibit this property.  
Figure \ref{fig:A} shows two of the learned transformation operators,
where
each $11 \times 11$ block corresponds to one column of \textit{A} and the
block's position in the figure corresponds to its pixel location in the original
image patch.  This can be viewed as an array of basis functions, 
each one
showing how intensity at a given pixel location influences the instantaneous change in pixel
intensity at all pixel locations (see Equation \ref{equ:derivative}). In this figure, each basis has become a
spatial differentiator. 
The bottom two rows of Figure \ref{fig:A} show each
of the operators being applied to an image patch. 
An animation of the full set
of learned affine operators applied to image patches can be found 
at \url{http://redwood.berkeley.edu/jwang/affine.html} and in the
supplemental material.

\begin{figure}
\begin{center}
(a) \parbox[b]{6cm}{\center Horizontal Translation\\\vspace{2mm}
\includegraphics[width=6cm]{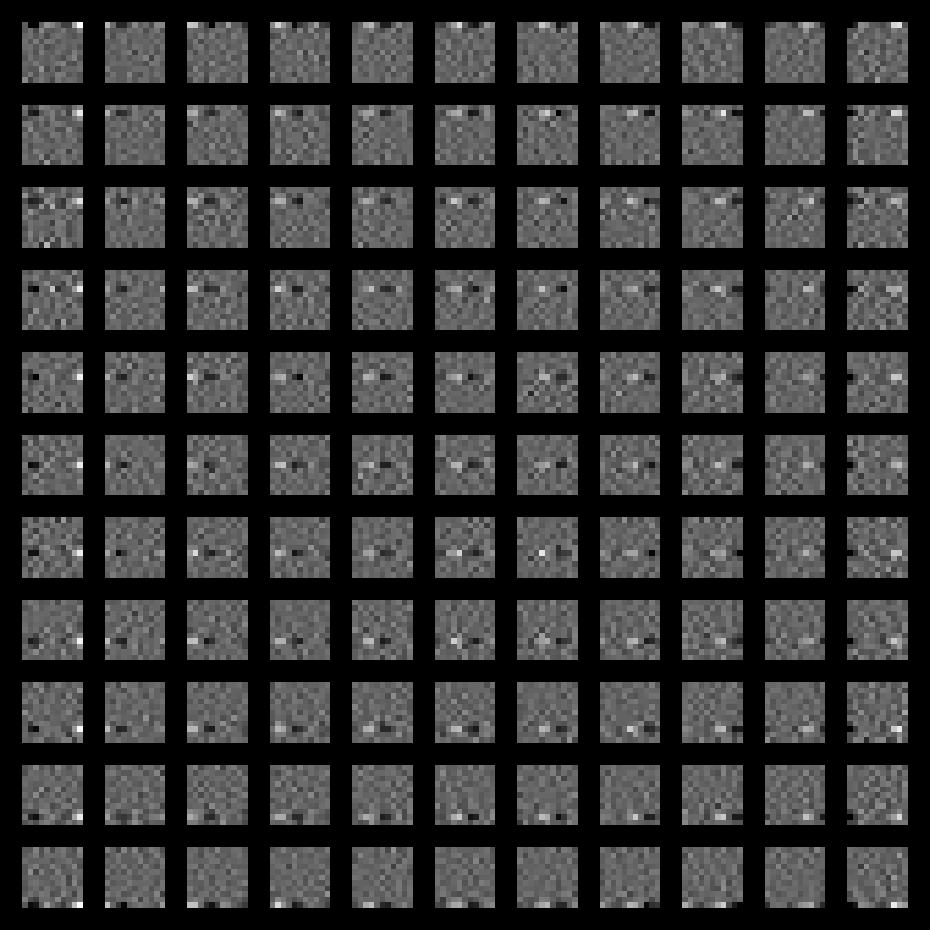}}
\ \ (b) \parbox[b]{6cm}{\center Rotation\\\vspace{2mm}
\includegraphics[width=6cm]{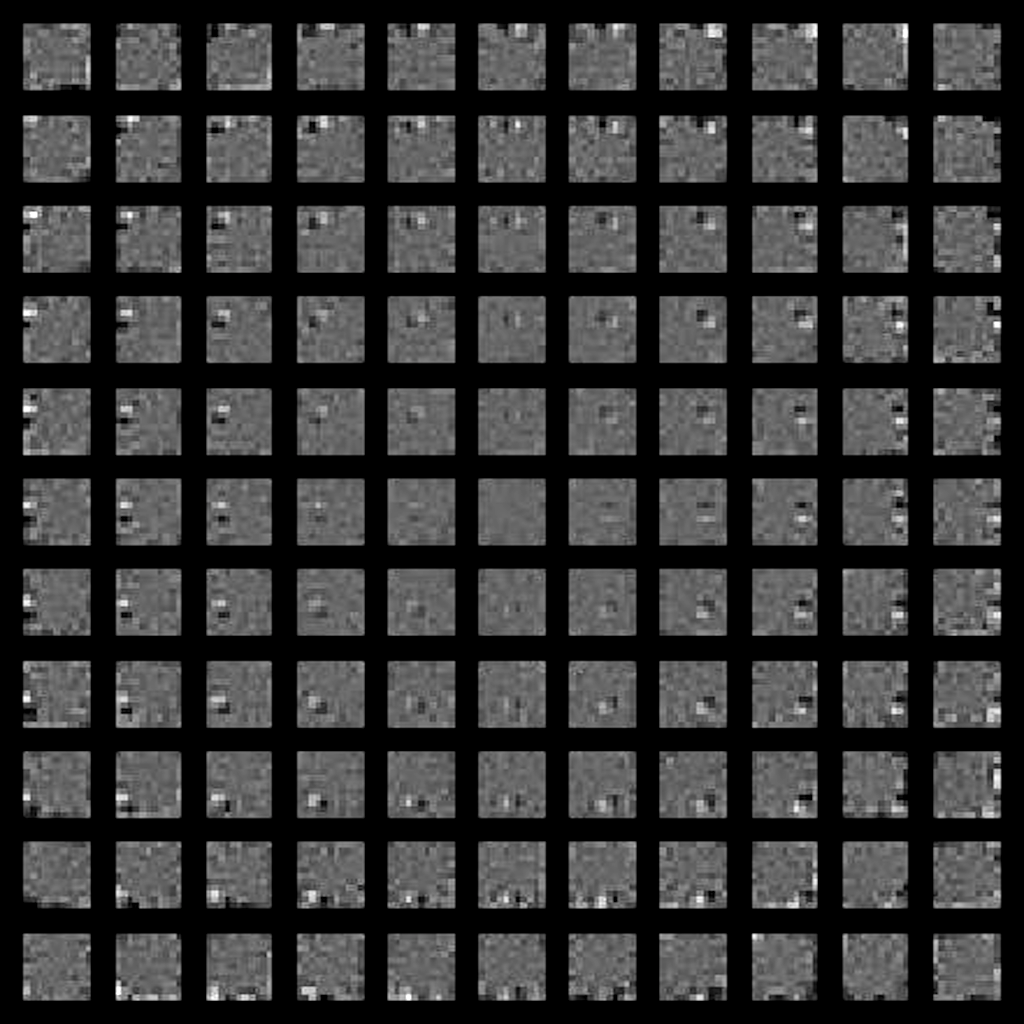}} \\
~\\
(c) \includegraphics[width=12cm,trim=0mm 0mm 0mm 6mm,clip]{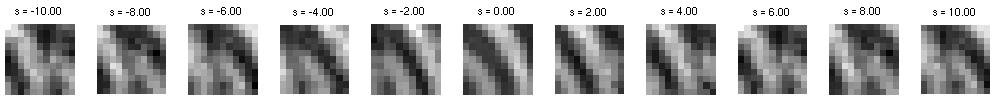} \\ ~ \\
(d) \includegraphics[width=12cm,trim=0mm 0mm 0mm 6mm,clip]{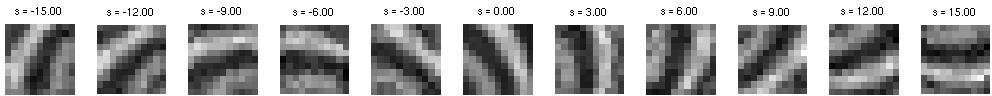}
\end{center}
\caption{The learned transformation operators that corresponds to horizontal translation
{\em(a)} and rotation {\em(b)}.  Each $11 \times 11$ block corresponds
to one column of $A$ and the
block's position in the figure corresponds to its pixel location in the original
image patch.  Each block therefore shows how intensity at one pixel location contributes to the instantaneous change in intensity at all other pixel locations.  Note that the blocks correspond to spatial derivatives in the direction of motion. 
Panels {\em(c)} and {\em(d)} show the translation and rotation operators, respectively, being applied to an image patch.}
\label{fig:A}
\end{figure}
\begin{figure}
\begin{center}
(a) \includegraphics[width=6cm]{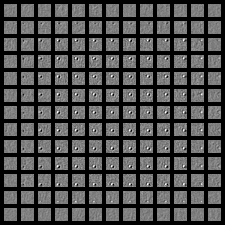}\ \ \ 
(b) \includegraphics[width=6cm]{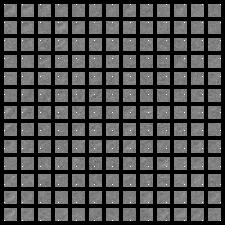} \\ ~ \\
(c) \includegraphics[width=6cm]{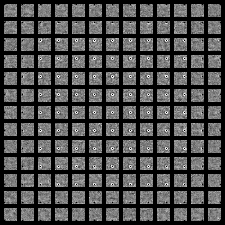}\ \ \ 
(d) \includegraphics[width=6cm]{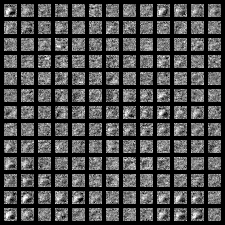}

\end{center}
\caption{Sample transformation operators from a set of 15 transformations trained in an unsupervised fashion on 13x13 pixel patches (including a 2 pixel wide buffer region) from natural video.  Each $13 \times 13$ block corresponds
to one column of $A$ and the
block's position in the figure corresponds to its pixel location in the original
image patch.  Each block therefore illustrates the influence a single pixel has on the entire image patch as the transformation is applied.  {\em (a)} full field translation operator, {\em (b)} full field intensity scaling, {\em (c)} full field contrast scaling, {\em (d)} unknown or difficult to interpret.
}
\label{fig:15 trans}
\end{figure}

\subsection{Learning Transformations from Natural Movies}  \label{sec time}
To explore the transformation statistics of natural images, we trained the algorithm on pairs of $17 \times 17$ pixel image patches cropped from 
consecutive frames from 
the same video dataset as in Sections \ref{section:infer}  and \ref{sec affine learn}.  
In order to allow the learned transformations to capture image features moving into and out of a patch from the surround, and to allow more direct comparison to motion compensation algorithms, the error function for inference and learning was
only applied to the central 9 $\times$ 9 region in each $17 \times 17$ patch.  Each patch can therefore be viewed as a 9 $\times$ 9 patch surrounded by a 4 pixel wide buffer region.  In the 15 transformation case a 2 pixel wide buffer region was used for computational reasons, so the 15 transformation case acts on 13 $\times$ 13 pixel patches with the reconstruction penalty applied on the central 9x9 region. 
Initialization was the same as in Section \ref{section:infer}.

Training was performed on a variety of models with different numbers of transformations.  For several of the models two of the operators were pre-specified to be whole-patch horizontal and vertical translation.  This was done since we expect that translation will be the predominant mode of transformation in natural video, and this allows the algorithm to focus on learning less obvious transformations contained in video with the remaining operators.  This belief is supported by the observation that several operators converge to full field translation when learning is unconstrained, as illustrated by the operator in Figure \ref{fig:15 trans}a from the 15 transformation case.  Prespecifying translation also provides a useful basis for comparing to existing motion compensation algorithms used in video compression, which are based on whole-patch translation.

The model case with the greatest variety of transformation operators consisted of 15 unconstrained transformations.  A selection of the learned $\mb A_k$ is shown in Figure \ref{fig:15 trans}.  
The learned transformation operators performed full field translations, intensity scaling, contrast scaling, spatially localized mixtures of the preceding 3 transformation types, and a number of transformations with no clear interpretation.  

\begin{figure}
\begin{center}
	\includegraphics[width=12cm]{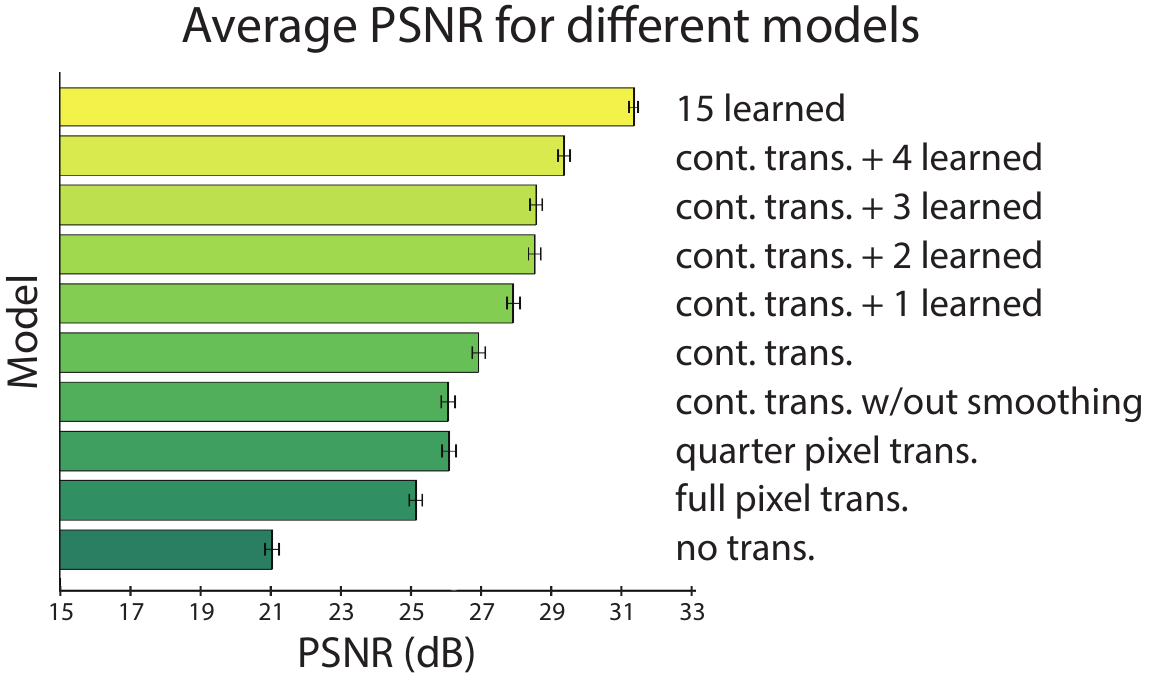}
\end{center}
\caption{More complex models allow for more accurate representation of inter-frame differences.  Bars show the PSNR of the reconstruction of a second frame via transformation of a first frame averaged over 1,000 pairs of frames from natural video, for a variety of model configurations.}
\label{fig:bar_psnr}
\end{figure}
To demonstrate the effectiveness of the learned transformations at capturing the interframe changes in natural video, the PSNR of the image reconstruction for 
1,000 pairs of $17 \times 17$ image patches extracted from consecutive frames 
was compared for all of the learned transformation models, as well as to standard motion compensation reconstructions.  The models compared were as follows:
\begin{enumerate}
 \item No transformation.  Frame $\mb x^{(t)}$ is compared to frame $\mb x^{(t+1)}$ without any transformation.
 \item Full pixel motion compensation.  The central $9 \times 9$ region of $\mb x^{(t+1)}$ is compared to the best matching $9 \times 9$ region in $\mb x^{(t)}$ with full pixel resolution.
 \item Quarter pixel motion compensation with bilinear interpolation.  The central $9 \times 9$ region of $\mb x^{(t+1)}$ is compared to the best matching $9 \times 9$ region in $\mb x^{(t)}$ with quarter pixel resolution.
 \item Continuous translation without smoothing. Only vertical and horizontal translation
operators are used in the model, but they are allowed to perform subpixel translations. 
 \item Continuous translation. Vertical and horizontal translation
operators are used in the model, and in addition adaptive smoothing is used. 
 \item Continuous translation plus learned operators. Additional transformation operators are randomly initialized and learned
in an unsupervised fashion.
 \item 15 learned transformation operators. Fifteen operators are randomly initialized and learned
in an unsupervised fashion.  No operators are hard coded to translation.
\end{enumerate}


As shown in Figure \ref{fig:bar_psnr} there is a steady increase in PSNR as the 
transformation models become more complex.  This suggests that as operators are added to 
the model they are learning transformations that are matched to progressively more complex 
frame-to-frame changes occurring in natural movies.  Note also that continuous translation with 
adaptive smoothing yields a substantial improvement over quarter pixel translation. This suggests 
that the addition of a smoothing operator itself could be useful even when employed with the 
standard motion compensation model.

%
%
%
%

\section{Discussion}

We have described and tested a method for learning Lie group operators from 
high dimensional time series data, specifically image sequences extracted from natural movies.  
This builds on previous work using Lie group operators to represent image transformations 
by making four key contributions:  1) an eigen-decomposition of the Lie group operator, 
which allows for computationally tractable learning; 2) an adaptive smoothing operator that 
reduces local minima during inference; 3) a mechanism for combining multiple non-commuting transformation operators; and 
4) a method for regularizing coefficients by manifold distance traversed.  The results obtained from both inferring transformation coefficients 
and learning operators from natural movies demonstrate the effectiveness of this method for representing transformations in complex, high-dimensional data.

In contrast to traditional video coding which uses a translation-based motion compensation algorithm, our method 
attempts to learn an optimal representation of image transformations from the statistics of the image sequences.  
The resulting model discovers additional transformations (beyond simple translation) to code natural video, yielding 
substantial improvements in frame prediction and reconstruction. The transformations learned on natural video include 
intensity scaling, contrast scaling, and spatially localized affine transformations, as well as full field translation.

The improved coding of inter-frame differences in natural video points to the potential utility of this algorithm for 
video compression.  For this purpose the additional cost of encoding the transformation coefficients ($\mu$ and $\sigma$) 
needs to be accounted for and weighed against the gains in PSNR of the predicted frame. This tradeoff between 
encoding and reconstruction cost is explored, and a rate-distortion analysis performed, in a separate paper \citep{DCC_11}.

Beyond video coding, the Lie group method could also be used to represent complex motion in natural movies for the purpose 
of neurophysiological or psychophysical experiments.  For example, neural responses could be correlated against the values 
of the inferred coefficients of the transformation operators.  Alternatively, synthetic video sequences could be generated using 
the learned transformations, and the sensitivity of neural responses, or an observer, could be measured with respect to changes 
in the transformation coefficients. In this way, the learned transformation operators could provide a more natural set of features 
for probing the representation of transformations in the brain.

Another extension of this method would be to learn the transformations in a latent variable representation of the image, rather 
than directly on pixels.  For example, one might first learn a dictionary to describe images using sparse coding 
\citep{Olshausen1996}, and then model transformations among the sparse coefficients in response to a natural movie.  
Alternatively, one might infer a 3D model from the 2D image and then learn the operators underlying the 3D transformations of 
objects in the environment, following the approach of \citep{bregler1998tracking}.
\bibliographystyle{plainnat}
\bibliography{neco}        

%
%
%


\appendix
 \renewcommand{\theequation}{A-\arabic{equation}}
  \setcounter{equation}{0}  
\section{Appendix - Degeneracy in U}
\label{u_rescale}

We decompose our transformation generator
\begin{equation}
A = V \Lambda V^{-1}
\end{equation}
where $\Lambda$ is diagonal.  We introduce another diagonal matrix $R$.  The diagonal of $R$ can be populated with any non-zero complex numbers, and the following equations will still hold:
\begin{eqnarray}
A & = & V \Lambda V^{-1} \\
 & = & V R R^{-1} \Lambda V^{-1} \\
 & = & V R \Lambda R^{-1} V^{-1} \\
 & = & \left(VR\right) \Lambda \left(VR\right)^{-1}
\end{eqnarray}

If we set
\begin{equation}
U = V R
\end{equation}
then
\begin{equation}
A = U \Lambda U^{-1}
\end{equation}
and $R$ represents a degeneracy in $U$.

We remove this degeneracy in $U$ by choosing $R$ so as to minimize the joint power after every learning step.  That is
\begin{eqnarray}
R & = & \argmin_R \sum_i \sum_j V_{ij}^2 R_{jj}^2 + \sum_i \sum_j \left(R^{-1}\right)_{jj}^2 \left(V^{-1}\right)_{ji}^2 \\
    & = & \argmin_R \sum_i \sum_j V_{ij}^2 R_{jj}^2 + \sum_i \sum_j \left(V^{-1}\right)_{ji}^2 \frac{1}{R_{jj}^2}
\end{eqnarray}
setting the derivative to $0$
\begin{eqnarray}
2 \sum_i V_{ij}^2 R_{jj} - 2 \sum_i \left(V^{-1}\right)_{ji}^2 \frac{1}{R_{jj}^3} & = & 0 \\
R_{jj} \sum_i V_{ij}^2 & = & \frac{1}{R_{jj}^3} \sum_i \left(V^{-1}\right)_{ji}^2 \\
R_{jj}^4 & = & \frac{
	\sum_i \left(V^{-1}\right)_{ji}^2}{
	\sum_i V_{ij}^2} \\
R_{jj} & = & \left[ \frac{
	\sum_i \left(V^{-1}\right)_{ji}^2}{
	\sum_i V_{ij}^2}  \right]^{\frac{1}{4}}
.
\end{eqnarray}

Practically this means that, after every learning step, we set
\begin{eqnarray}
R_{jj} & = & \left[ \frac{
	\sum_i \left(U^{-1}\right)_{ji}^2}{
	\sum_i U_{ij}^2}  \right]^{\frac{1}{4}}
\end{eqnarray}
and then set
\begin{eqnarray}
U_{new} = UR
.
\end{eqnarray}

 \renewcommand{\theequation}{B-\arabic{equation}}
  \setcounter{equation}{0}  
  
\section{Appendix - Derivatives}

Let
\begin{equation}
 \varepsilon = \sum_n ( Y_n - U e^{\mu \Lambda} e^{\frac{1}{2} \Lambda^2
\sigma^2}U^{-1} X_n )^2
\end{equation}

\subsection{ Derivative for inference with one operator }

The learning gradient with respect to $\mu$ is 
\begin{equation}
\begin{split}
 \frac{\partial \varepsilon}{\partial \mu } &= -\sum_n 2 err(n) U
\frac{ \partial e^{\mu \Lambda} }{\partial \mu} e^{\frac{1}{2} \Lambda^2
\sigma^2}U^{-1} X_n \\
 & = -\sum_n 2 err(n) U
\Lambda e^{\mu \Lambda}  e^{\frac{1}{2} \Lambda^2 \sigma^2}U^{-1} X_n \\  
\end{split}
\end{equation}

where $err(n)$ is the reconstruction error of the n$^{th}$ sample

\begin{equation}
err(n) = Y_n - U e^{\mu \Lambda} e^{\frac{1}{2} \Lambda^2 \sigma^2}U^{-1}
X_n
\end{equation}

Similarly, the learning gradient with respect to $\sigma$ is 
\begin{equation}
\begin{split}
 \frac{\partial \varepsilon}{\partial \sigma } &= -\sum_n 2 err(n) U e^{\mu
\Lambda} \sigma \Lambda^2 e^{\frac{1}{2} \Lambda^2 \sigma^2}U^{-1} X_n \\ 
\end{split}
\end{equation}

\subsection{ Derivative for learning with one operator }

The learning gradient with respect to $\Lambda$ is 
\begin{equation}
\begin{split}
 \frac{\partial \varepsilon}{\partial \Lambda } &= -\sum_n 2 err(n) U
\frac{ \partial \left[e^{\mu \Lambda} e^{\frac{1}{2} \Lambda^2
\sigma^2} \right]}{ \partial \Lambda} U^{-1} X_n \\
&= -\sum_n 2 err(n) U ( \mu
e^{\mu \Lambda} + \sigma^2 \Lambda e^{\frac{1}{2} \Lambda^2
\sigma^2} ) \ U^{-1} X_n \\
\end{split}
\end{equation}
The learning gradient with respect to U is 
\begin{equation}
\begin{split}
 \frac{\partial \varepsilon}{\partial U } =& -\sum_n 2 err(n) \frac{\partial
U}{\partial U} e^{\mu \Lambda} e^{\frac{1}{2} \Lambda^2
\sigma^2} U^{-1} X_n \\
& -\sum_n 2 err(n) U e^{\mu \Lambda} e^{\frac{1}{2} \Lambda^2
\sigma^2} \frac{ \partial U^{-1} }{\partial U} X_n  
\end{split}
\end{equation}
Recall that 
\begin{equation}
\frac{ d U^{-1} }{U} = - U^{-1} \frac{ d U } {d U} U^{-1}.
\end{equation}
The learning gradient is therefore
\begin{equation}
\begin{split}
 \frac{\partial \varepsilon}{\partial U } =& -\sum_n 2 err(n) e^{\mu \Lambda}
 e^{\frac{1}{2} \Lambda^2 \sigma^2} U^{-1} X_n \\
 & +\sum_n 2 err(n) U e^{\mu \Lambda} e^{\frac{1}{2} \Lambda^2
 \sigma^2} U^{-2} X_n  
\end{split}
\end{equation}

\subsection{ Derivative for Complex Variables }
To accommodate for the complex variables U and $\Lambda$, we rewrite our
objective function as 
\begin{equation}
\varepsilon = \sum_n err(n)^T \overline{err(n)}
\end{equation}
where $\overline{err(n)}$ denotes the complex conjugate. The
derivative of this error function with respect to any complex variable can be
then broken into the derivative with respect to the real and and imaginary parts.
This results in
%
\begin{equation}
\begin{split}
\frac{\partial \varepsilon}{\partial \Real\left \{\Lambda\right\}} &= 2 \Real
\left \{ \sum_n err(n)^T \overline {\left ( \frac{ \partial err(n) }{\partial
\Lambda} \right ) } \right \} \\
 \frac{\partial \varepsilon}{\partial \Imag \left \{ \Lambda\right \}}  &= -2
\Imag \left \{ \sum_n err(n)^T \overline {\left ( \frac{ \partial err(n)
}{\partial \Lambda} \right ) } \right \}
,
\end{split}
\end{equation}
(similarly for $U$).
%
%
\subsection{ Derivatives for Manifold Penalty }

We have a model
\begin{equation}
\dot{I} = A I
\end{equation}
with solution
\begin{equation}
I_1 = e^{As}I_0
.
\end{equation}

We want to find and minimize the distance traveled by the image patch $I_0$ to
$I_1$ under the transformation operator A. The total distance is 
\begin{equation}
d = \int_{t=0}^s || \dot{I} ||_2^1 dt
.
\end{equation}
This then gives the following,
\begin{equation}
\begin{split}
  d &= \int_{t=0}^s ||A x^{(t)}||_2^1 dt \\
 & = \int_{t=0}^s || A e^{At} I_0 ||_2^1 dt \\
 & = \int_{t=0}^s \sqrt{ (A e^{At} I_0)^T (A e^{At} I_0) } dt \\
 & = \int_{t=0}^s \sqrt{ I_0^T (e^{At})^T A^T A e^{At} I_0 } dt \\
\end{split}
.
\end{equation}
We don't know how to solve this analytically. Instead we make the approximation 
\begin{equation}
\begin{split}
   \int_{t=0}^s || \dot{I} ||_2 dt & \approx \left|s\right| ||A I_{\frac{s}{2}}||_2 \\
   & = \left|s\right| || A e^{A \frac{s}{2} } I_0 ||_2
\end{split}
,
\end{equation}
with derivative
\begin{equation}
\begin{split}
\frac{\partial }{\partial s} \left[\left|s\right| || A e^{A \frac{s}{2} } I_0 ||_2 \right] &=
  B + 2sB
\end{split}
,
\end{equation}
where $ B = I_0^T (e^{A \frac{s}{2} })^T A^T A e^{A \frac{s}{2} } I_0 $

\end{document}